\documentclass[conference]{IEEEtran}
\IEEEoverridecommandlockouts
\usepackage[colorlinks,citecolor=blue,urlcolor=blue,bookmarks=false,hypertexnames=true]{hyperref} 
\usepackage{cite}
\usepackage{amsmath,amssymb,amsfonts}
\usepackage{subcaption}
\usepackage{graphicx}
\usepackage{textcomp}
\usepackage{xcolor}
\usepackage{svg}
\usepackage{float}
\usepackage{amsmath}
\usepackage{multicol}
\usepackage{array}
\usepackage{mwe}
\usepackage[noend]{algpseudocode}
\usepackage[ruled, lined, longend]{algorithm2e}
\usepackage{multirow}
\def\BibTeX{{\rm B\kern-.05em{\sc i\kern-.025em b}\kern-.08em
    T\kern-.1667em\lower.7ex\hbox{E}\kern-.125emX}}
\begin{document}

\title{\huge MicroNAS: Zero-Shot Neural Architecture Search for MCUs}

\author{\IEEEauthorblockN{Ye Qiao, Haocheng Xu, Yifan Zhang, Sitao Huang}
\IEEEauthorblockA{\textit{Department of Electrical Engineering and Computer Science} \\
\textit{University of California, Irvine,} Irvine,  California, USA \\
\{yeq6, haochx5, yifanz58, sitaoh\}@uci.edu}
}

\maketitle

\begin{abstract}

Neural Architecture Search (NAS) effectively discovers new Convolutional Neural Network (CNN) architectures, particularly for accuracy optimization. However, prior approaches often require resource-intensive training on super networks or extensive architecture evaluations, limiting practical applications. To address these challenges, we propose MicroNAS, a hardware-aware zero-shot NAS framework designed for microcontroller units (MCUs) in edge computing. MicroNAS considers target hardware optimality during the search, utilizing specialized performance indicators to identify optimal neural architectures without heavy computational costs. Compared to previous works, MicroNAS achieves up to $1104\times$ improvement in search efficiency and discovers models with over $3.23\times$ faster MCU inference while maintaining similar accuracy.

\end{abstract}

\section{Introduction}
Manually designing high-accuracy or computationally efficient Convolutional Neural Network (CNN) topologies in computer vision, speech recognition, and object detection demands significant time, expertise, and resources. These models are often too large for deployment in edge environments like microcontroller units (MCUs) due to resource limitations. To automate model design within constraints, neural architecture search (NAS) is essential \cite{lin2020mcunet}. While NAS has evolved, most approaches still face time-consuming training and evaluation \cite{muNAS}. In this context, we introduce MicroNAS, an efficient zero-shot NAS framework tailored for MCUs. MicroNAS enables effective architecture search for edge devices without excessive computational costs, facilitating practical NAS deployment in edge computing. 

The main contributions of this work are: \textbf{MicroNAS Proposal:} We introduce MicroNAS, a novel zero-shot NAS framework for optimal CNN architectures in MCU-based inference.
\textbf{Hybrid Objective Function:} Our proposed objective function, combining neural tangent kernel spectrum, linear region count, and hardware proxies, significantly enhances NAS quality for MCUs.
\textbf{Hardware-Aware Pruning-Based Search Algorithm:} We propose an innovative pruning-based search algorithm to improve search efficiency under resource constraints. \textbf{Analysis Results:} Using trainless proxies on NAS-Bench-201 space, we achieve up to $1104\times$ improvement in search efficiency compared to prior works, discovering models with over $3.23\times$ faster MCU inference while maintaining similar accuracy.
This work marks a substantial stride toward efficient NAS solutions in edge environments.

 \begin{figure*}[ht]
    \centering 
    \vspace{-7mm}
    \includegraphics[width=1\textwidth]{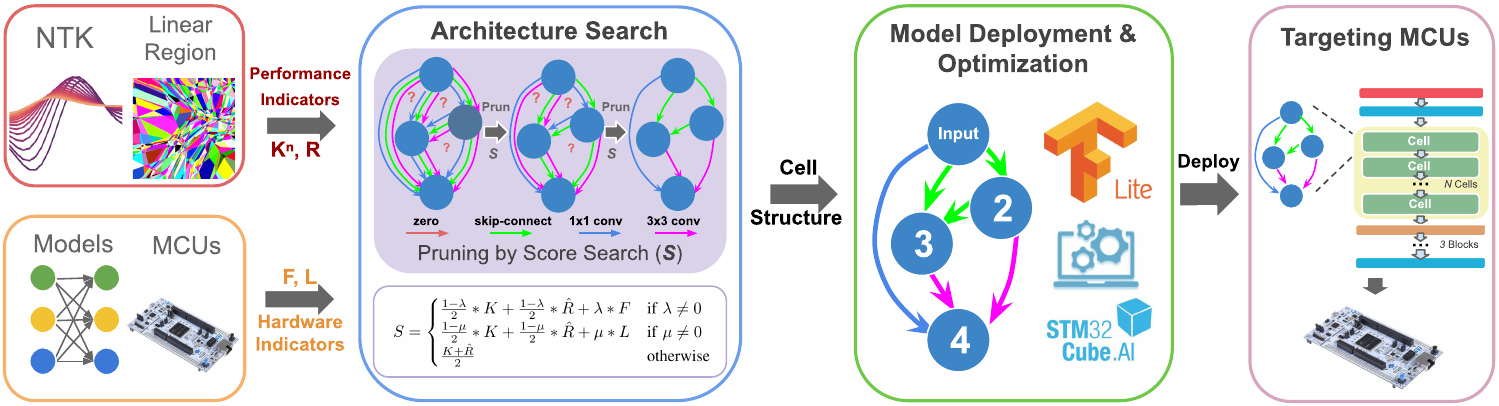}
    \caption{Overview of Proposed MicroNAS Workflow}
    \vspace{-4mm}
    \label{fig:framwork}
\end{figure*}
\section{Proposed MicroNAS Framework}

To eliminate the need for training and evaluation during the architecture search process, we employ key indicators that capture the trainability\cite{chen2021neural}, expressivity\cite{lin2021zennas}, and hardware performance of neural networks. We incorporate multiple zero-cost proxies and generalize the search to discover the optimal cell structure for MCUs. A cell-based search space defines each architecture as a directed acyclic graph (DAG), with nodes representing feature maps and edges corresponding to operations as shown in Fig. \ref{fig:framwork}.

\vspace{-1mm}
\subsection{Performance Indicators}
\subsubsection{Spectrum of Neural Tangent Kernel}
\begin{figure}[ht]
\vspace{-3mm}
    \centering
    \begin{subfigure}[b]{.48\columnwidth}
        \centering
        \includegraphics[width=\linewidth]{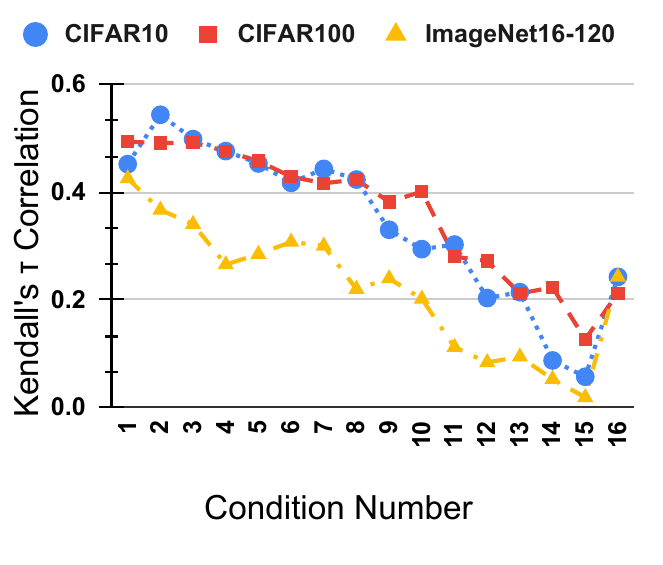}
        \vspace{-7mm}
        \caption{Kendell's $\tau$ vs. $K_i$}
        \label{ntkvst}
    \end{subfigure}
    \hfill
    \begin{subfigure}[b]{.48\columnwidth}
        \centering
        \includegraphics[width=\linewidth]{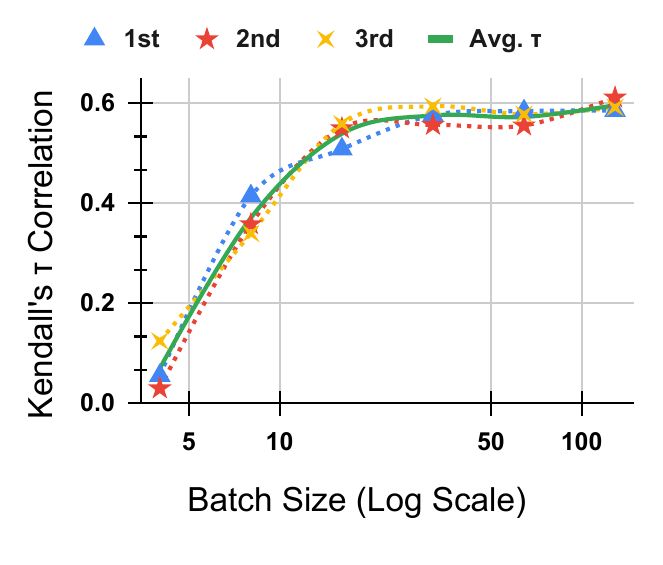}
        \vspace{-7mm}
        \caption{Kendall-$\tau$ vs. Batch Size}
        \label{batchvst}
    \end{subfigure}
    \vspace{-1mm}
    \caption{Kendall-$\tau$ vs Batch Size, Condition Number $K_i$ }
    \vspace{-5mm}
    \label{fig:tau}
\end{figure}

The neural tangent kernel (NTK) is a mathematical construct used to analyze the behavior and properties of neural networks \cite{xiao2020disentangling}. Indeed, the trainability of a neural network, which refers to its convergence and generalization ability, is a crucial aspect of zero-shot NAS. The NTK's condition number relies on a single mini-batch, and the chosen batch size affects NTK spectrum consistency, influencing search outcomes. Investigating Kendall-$\tau$ correlation against logarithmic batch size scales, Fig. \ref{batchvst} reveals an optimal batch size range of 16 to 32. Increasing beyond 32 to 128 doesn't notably alter Kendall-$\tau$ correlation but significantly escalates search costs. Our experiments empirically evaluate and adopt a batch size of 32 for optimal search.

\subsubsection{Linear Region Count} In our study, we assess the expressivity of a simple CNN with each layer containing a single convolutional operator followed by the ReLU activation function. The ReLU's piecewise linearity allows the network's input space to be divided into distinct linear regions (LR) \cite{xiong2020number}. Each LR is associated with a set of affine parameters, and the network's expressivity is determined by the number of these linear regions it can separate.

\vspace{-1mm}
\subsection{Hardware Indicators}
The hardware-aware search process focuses on low-power edge microcontroller units, emphasizing effective management of computation costs, processing latency, and memory usage. We prioritize inference latency as it directly influences real-world processing time. To address this, we integrate Floating Point Operations (FLOPs) estimation ($F$) and hardware latency modeling ($L$) into the architecture search for sampled models. Our approach includes tunable weight factors for precise control over the contributions of $F$ and $L$ during the search.
\subsubsection{The Floating Point Operations (FLOPs)}

FLOPs count is a crucial indicator of deep learning model complexity, reflecting computing time in the target environment regardless of hardware specialization. Our estimation considers various layer operations. While our experiments show a positive correlation between FLOPs count and model accuracy, it's important to note that FLOPs alone don't represent absolute accuracy or real-world hardware performance due to redundancy and topological differences in convolutional neural networks. Therefore, we introduce estimated model latency as an additional hardware indicator for a more comprehensive model assessment.
\subsubsection{Estimated Latency}

To address this, we developed a custom latency estimator to accurately model inference latency on the target MCUs based on provided cell structures. The approach involves profiling each operation individually within the search space and generating a reference lookup table. Specific details of the secondary stage of the model structure, including the number of cells and input/output channels for each cell, are gathered. Finally, constant hardware latency overhead is profiled and incorporated into the overall latency estimation. Our latency model was validated as accurate, reliable, and simple in further experiments with the integration of our hardware-aware pruning-based search.

\begin{table}[ht]
\vspace{-2mm}
\caption{Results on CIFAR-10}
\centering
\resizebox{\columnwidth}{!}{%
\begin{tabular}{cccccc}
\hline
NAS Frameworks & FLOPs (M) & Params (M) & Speedup& Search Time& ACC \\ \hline
$\mu$NAS\cite{muNAS}   & -              & 0.014         & -              & 552            & 86.49                 \\
TE-NAS\cite{chen2021neural}  & 188.66         & 1.317       & 1             & 0.43           & 93.78      \\
Ours & \textbf{51.04} & \textbf{0.372}  & \textbf{3.23$\times$}  & \textbf{0.43} & \textbf{93.88} \\ \hline
\end{tabular}%
}


\label{table2}
\end{table}
\vspace{-2mm}



\section{Experiment and Results Analysis}

We utilized a STM32 NUCLEO-F746ZG board and evaluate our MicroNAS on NAS-Bench-201 \cite{dong2020nasbench201}.

In the search, MicroNAS adapts FLOPs and latency indicator weights, consistently discovering highly efficient models across various constraints with minimal performance degradation. Even without hardware constraints, our baseline result surpasses state-of-the-art TE-NAS \cite{chen2021neural}. Our hardware-aware strategy provides a latency advantage of $1.59\times$ to $3.23\times$ with negligible performance trade-offs. Notably, our latency-guided search outperforms TE-NAS \cite{chen2021neural}, halving inference time on the test MCU. Compared to $\mu$NAS \cite{muNAS}, another NAS for edge hardware, MicroNAS finds significantly better-performing models with approximately $1104\times$ efficiency in search time (reported in GPU hours) and $6.2\%$ better performance. The latency-guided search demonstrates superior and more balanced performance than the FLOPs-guided search, attributed to MCU-specific bias in our latency modeling. Both approaches are valuable for MCU platforms, with fine-grained latency estimation resulting in superior search outcomes.

\section{Conclusion}


This paper introduces MicroNAS, a framework that combines neural network analysis and target MCU hardware constraints for identifying optimal architectures with minimal performance trade-offs and low training and evaluation costs. The proposed MCU latency estimation model accurately predicts real-world inference latency on MCUs and has potential applicability to other edge devices. MicroNAS achieves a substantial search efficiency improvement, surpassing previous works by $1104\times$, discovering models with more than $3.23\times$ faster MCU inference while maintaining comparable accuracy. Future experiments will incorporate peak memory usage modeling of MCUs to guide the search and enhance the MicroNAS framework further.





\bibliography{refernces.bib}{}

\bibliographystyle{IEEEtran}

\end{document}